\PassOptionsToPackage{table}{xcolor}
\documentclass[10pt,twocolumn,letterpaper]{article}

\usepackage{iccv}
\usepackage{times}
\usepackage{epsfig}
\usepackage{graphicx}
\usepackage{amsmath}
\usepackage{amssymb}
\usepackage{mathrsfs}
\usepackage{caption}
\usepackage{graphbox}
\usepackage{diagbox}
\usepackage{float}
\usepackage[dvipsnames]{xcolor}
\usepackage[ruled,linesnumbered]{algorithm2e}
\usepackage{booktabs}
\usepackage{marvosym}

\usepackage[pagebackref=true,breaklinks=true,letterpaper=true,colorlinks,bookmarks=false]{hyperref}
\usepackage[table]{xcolor}
\usepackage[capitalize]{cleveref}
\Crefname{section}{Section}{Sections}
\crefname{section}{Sec.}{Secs.}
\Crefname{align}{Equation}{Equations}
\crefname{align}{Eq.}{Eqs.}
\Crefname{equation}{Equation}{Equations}
\crefname{equation}{Eq.}{Eqs.}
\Crefname{figure}{Figure}{Figures}
\crefname{figure}{Fig.}{Figs.}
\Crefname{table}{Table}{Tables}
\crefname{table}{Tab.}{Tabs.}

\iccvfinalcopy 

\def\iccvPaperID{****} 

\ificcvfinal\fi
\begin{document}

\title{Devil in the Number: Towards Robust Multi-modality Data Filter}

\author{
    Yichen Xu\textsuperscript{1} \quad
    Zihan Xu\textsuperscript{2} \quad
    Wenhao Chai\textsuperscript{3$\heartsuit$} \quad
    Zhonghan Zhao\textsuperscript{2} \quad
    Enxin Song\textsuperscript{2} \quad
    Gaoang Wang\textsuperscript{2\Letter}\\
    [2mm]
    \textsuperscript{1}~Fudan University \quad \textsuperscript{2}~Zhejiang University \quad \textsuperscript{3}~University of Washington \\
    [2mm]
    \normalsize{ 
    \textsuperscript{$\heartsuit$}~Project lead \quad 
    \textsuperscript{\Letter}~Corresponding author}
}

\maketitle

\begin{abstract}

In order to appropriately filter multi-modality data sets on a web-scale, it becomes crucial to employ suitable filtering methods to boost performance and reduce training costs. For instance, LAION papers employs the CLIP score filter to select data with CLIP scores surpassing a certain threshold. On the other hand, T-MARS achieves high-quality data filtering by detecting and masking text within images and then filtering by CLIP score. Through analyzing the dataset, we observe a significant proportion of redundant information, such as numbers, present in the textual content. Our experiments on a subset of the data unveil the profound impact of these redundant elements on the CLIP scores. A logical approach would involve reevaluating the CLIP scores after eliminating these influences. Experimentally, our text-based CLIP filter outperforms the top-ranked method on the ``small scale" of DataComp (a data filtering benchmark) on  ImageNet distribution shifts, achieving a 3.6\% performance improvement. The results also demonstrate that our proposed text-masked filter outperforms the original CLIP score filter when selecting the top 40\% of the data. The impact of numbers on CLIP and their handling provide valuable insights for improving the effectiveness of CLIP training, including language rewrite techniques.

\end{abstract}

\section{Introduction}

As the scope of training datasets expands, it becomes crucial to ensure the presence of high-quality data for effective training. CLIP~\cite{radford2021clip}, as a vision-language model, also relies on high-quality text-image pairs. Recently, the initiative by DataComp~\cite{gadre2023datacomp} has further propelled efforts in this direction.
In terms of enhancing the CLIP score filter, Maini~\etal~\cite{maini2023t} proposed the T-MARS method (Text-Masking and Re-Scoring), which involves masking images. This inspires us to explore text masking as well. Through our exploration of text masking, we find the influence of redundant information such as the \textbf{number}.

We propose a novel filter named the text-masked CLIP method, which leverages the CLIP score. Specifically, we remove text sections containing numbers and bracketed content from consideration, subsequently calculating the CLIP similarity. Data is then filtered based on a predetermined proportion. This approach combines the given CLIP score filter with our text-masked CLIP score filter. Intriguingly, during our experimentation, we observe counter-intuitive behaviors exhibited by the CLIP model.

\begin{figure}[t]
    \centering
    \includegraphics[width=1\linewidth]{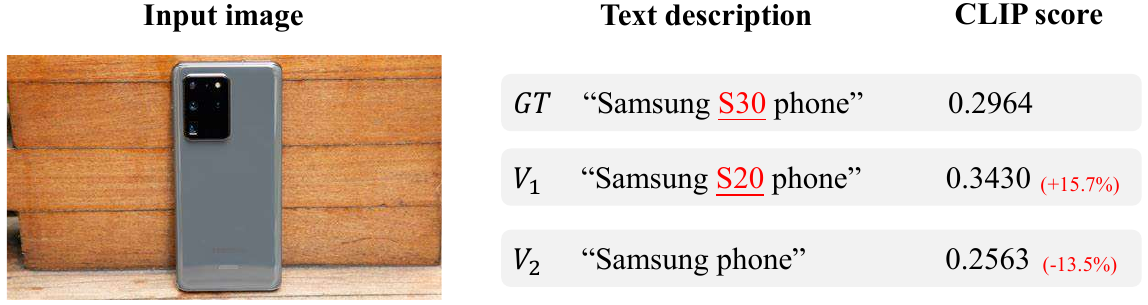} 
    \caption{The ambiguity of the CLIP model in the \textbf{number}. $GT$ refers to the ground truth text while $V_1$ and $V_2$ refer to two fake variation. We observe that the wrong text is given the highest score, the correct text that lacks numeric information is given the lowest score, and the correct text receives a mediocre score. It shows that the CLIP model is not robust regarding the \textbf{number}.}
    \label{fig:intro}
    \vspace{-10pt}
\end{figure}

Figure~\ref{fig:intro} demonstrates instances where the CLIP model occasionally fails to match image-text pairs accurately. Further analysis revealed that redundant information, such as numbers and bracketed contents, notably impacts the CLIP similarity score for image-text pairs. We hypothesize that some high-quality data might be contaminated by redundant information, resulting in a lower CLIP score than its actual quality warrants. Our experiments confirm this hypothesis, leading us to propose a new text-masked CLIP filter. This filter aims to identify and handle such data, thereby enhancing the robustness of our filter. 

Our main contributions are:
\begin{itemize}
    \vspace{-5pt}
    \item We observe that there is a large proportion of numbers in web-scale dataset, and some high-quality data can also manifest as low CLIP scores.
    \vspace{-5pt}
    \item By removing the bias introduced by redundant information like numbers, we can make the filtering method more robust. This is also an important consideration when optimizing data through rewriting techniques.
\end{itemize}

\section{Related Works}

\paragraph{Basic filtering.} Many simple filtering operations are considered, inspired by Schuhmann ~\etal ~\cite{schuhmann2021laion} and Byeon ~\etal ~\cite{kakaobrain2022coyo-700m}: filtering by language (English captions, using either fasttext ~\cite{joulin2016bag} or cld3 ~\cite{cld3}); filtering by caption length (over two words and five characters); and filtering by image size (smaller dimension above 200 pixels and aspect ratio below three). The combination of language and caption length filtering, as well as the combination of language, caption length, and image size filtering, is also experimented with. Unless otherwise specified, ``basic" refers to fasttext English, caption length, and image size filtering.

\vspace{-10pt}
\paragraph{CLIP score and LAION filtering.} CLIP score filtering (also employed by LAION) is implemented based on cosine similarity scores between CLIP image and text embeddings above a threshold.
A range of thresholds and two OpenAI CLIP models are investigated for computing the scores: the ViT-B/32 model (as in LAION) and the larger ViT-L/14. Combining CLIP score thresholds and cld3 English filtering is also done to reproduce the LAION-2B filtering scheme. 

\vspace{-10pt}
\paragraph{Text-based filtering.} Examples that contain text overlapping with ImageNet class names are selected, which serve as proxies for relevance to downstream tasks. Specifically, English captions (according to fasttext) that contain words from ImageNet-21K or ImageNet-1K ~\cite{5206848} class synsets are selected. 

\vspace{-10pt}
\paragraph{Image-based filtering.} A subset of examples whose visual content overlaps with ImageNet classes are selected. After applying English language (fasttext) and caption length filtering, the image embeddings extracted by the OpenAI ViT-L/14 model for each image are clustered into 100K groups using Faiss ~\cite{johnson2019billion}. The nearest neighbor group for every ImageNet training example is then found, and examples of these groups are kept. This procedure is applied using ImageNet-21K (14M images) or ImageNet-1K (1.2M images), forming two subsets.

\vspace{-10pt}
\paragraph{T-MARS filtering.} Recently, Maini ~\etal~\cite{maini2023t}.  introduced a filtering approach known as T-MARS (Text Masking and Re-Scoring). T-MARS employs the masking technique to conceal the text in images, followed by the computation of similarity scores between the masked image and its corresponding caption using a pretrained CLIP model. By excluding images with low masked similarity scores, the dataset can be curated effectively. Through experimental evaluation, T-MARS outperforms the previously top-ranked method on the ``medium scale" of DataComp.

\section{Method}
\definecolor{EX}{RGB}{244,177,131}
\definecolor{KP}{RGB}{47,85,151}

\begin{figure}
    \centering
    \includegraphics[width=1\linewidth]{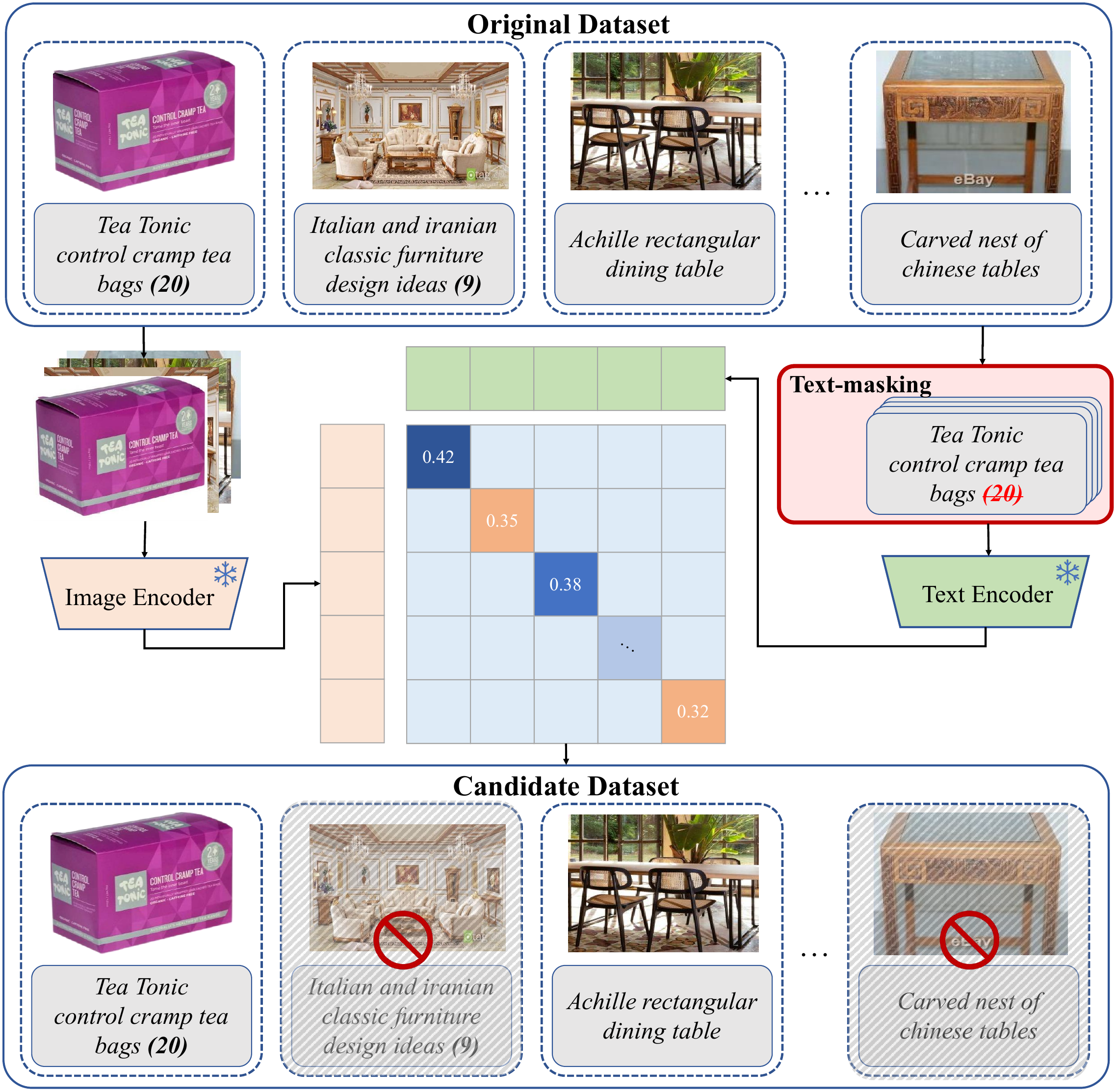}
    \caption*{
    \parbox[c][8pt][l]{8pt}{\colorbox{EX}{}}excluded\quad
    \parbox[c][8pt][l]{8pt}{\colorbox{KP}{}}kept\quad
    }
    \caption{A brief illustration of our text-masking filtering method with ratio 30\%. We perform text-masking on the texts and assess the CLIP similarities of the processed data. Then, we filter the original dataset based on the CLIP similarities by a given ratio to get the candidate dataset.}
    \label{fig:method}
\end{figure}

Numerical data can potentially mislead the CLIP model. Although not always redundant, the presence of numbers may act as a contaminant, causing some data to receive lower CLIP scores and be filtered out. Therefore, we propose removing numbers from texts to ensure a fairer comparison where all data is equally unaffected by numerical factors.

The proposed filter operates on text masking, which involves removing texts containing numbers and bracketed content to mitigate the impact of redundant or incorrect information. We then re-evaluate the CLIP L/14 similarity, taking into account the potential influence of numbers and bracketed contents on the CLIP score. By applying this mask, we can enhance the CLIP scores of high-quality data while reducing the scores of low-quality data. However, this approach may inadvertently lead to the loss of some high-quality data due to the side effects of the masking process.

As is shown in Algorithm~\ref{algorithm}, our approach shares similarities with the T-MARS method. However, instead of processing images as in T-MARS, we focus on processing texts containing numbers and bracketed contents. 

There are two main steps in our method:

\paragraph{Text masking.} In each text, we delete numbers and words with numbers, wiping out the bracketed contents. To illustrate, in Figure 1, we remove ``S30",  ``S20",  ``(View 18 of 50)",  and ``(20)". ``(View 18 of 50)" and ``(20)". 

\paragraph{Computing CLIP similarity and filtering.} Subsequently, we employ the ViT-L/14 laion2b\_s32b\_b82k pre-trained model to compute the cosine similarity between each image and text, followed by filtering the data based on the computed similarity.

\begin{algorithm}[t]
\caption{Text-masked CLIP filter}\label{algorithm}
\KwData{Dataset $\mathcal D$=$\{(i, t)\}^n$, CLIP Vision Encoder \emph{l} , text-masker \emph{m}}, fraction $\alpha$
\KwResult{filtered dataset $\mathcal D^*$}
\For{$k=0, \cdots, n-1$}{$\tilde {t_k}=m(t_k)$\\
$s_k=l(i_k, \tilde t_k)$}

threshold=the CLIP score of the [$n\alpha$] $th$ largest $s_k$

return \{$\mathcal D^*=(i_k, t_k)|s_k\geq $threshold\}
\end{algorithm}
\section{Experiments}
In the experimental section, we demonstrate the superiority of our method compared to the CLIP score filter with proportional representation. Furthermore, we validate the changes in data composition before and after filtration. We analyze the distribution of CLIP scores obtained from the filtered data, providing evidence for the introduction of low-quality data. Our hypothesis is strongly supported by the results obtained.

\begin{figure}
    \centering
\includegraphics[width=\linewidth]{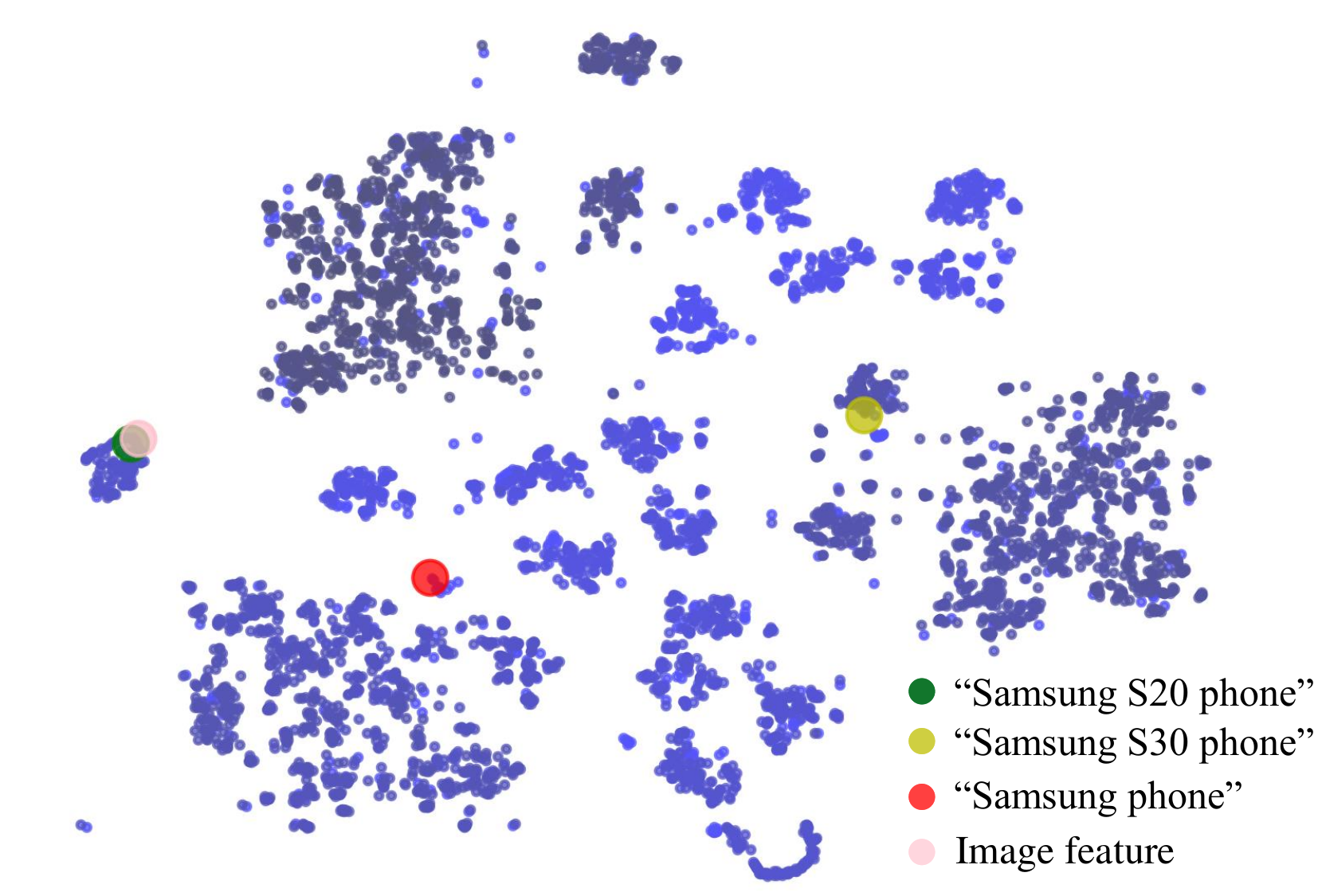}
    \caption{As an interpretation of Figure \ref{fig:intro}, the depicted visualization represents the distribution of text features from ``Samsung S1 phone" to ``Samsung S5000 phone". The darkness of the colors indicates the magnitude of the corresponding numerical values. Specifically, the red, green, and yellow dots represent the text features of ``Samsung phone", ``Samsung S20 phone", and ``Samsung S30 phone" respectively. The pink dot represents the image feature.}
    \label{fig:distribution}
\end{figure}

\vspace{-10pt}
\paragraph{A study conducted a simplified experiment on the Figure \ref{fig:intro}.} By utilizing the tokenizer and encoder provided by open-CLIP, both images and texts were transformed into features within the same space. Based on the observations from Figure \ref{fig:distribution}, it is evident that the ground truth text feature significantly deviates from the text features derived under scenarios involving erroneous or incomplete information, while also being distant from the image feature. The features obtained from different encoded numbers exhibit clear discreteness. Additionally, there are instances where data with substantially different numerical values yield relatively similar text features. This observation confirms that numerical values have the potential to introduce data contamination, resulting in substantial deviations in the feature space and consequently reducing CLIP scores.

\begin{table}[t]
    \centering
    \begin{tabular}{@{}c|c@{}}
        \toprule
        Filter  & w. num~(\%) \\
        \midrule
             CLIP 30\%& 48.6\\
             text-masked CLIP 30\%&45.1 \\
             CLIP 40\%& 47.6 \\
             text-masked CLIP 40\%&44.8 \\
        \bottomrule
        
    \end{tabular}
    \caption{Presence of number before and after filtering. }
    \label{tab:performance}
\end{table}

\vspace{-10pt}
\paragraph{Analysis of data composition in presence of number before and after filtering.} While filtering,the proportion of texts without numbers increases,as the statistics in ~\ref{tab:performance} shows. Our filtering method reduces the number of text-image pairs containing numbers, which is more evident in 40\% of cases. The statistical results show that our filter reduces the number of text-image pairs containing numbers compared to the CLIP filter. This will lead to the loss of some high-quality data, but expanding the filtering scale can ensure that the data quality is not too bad, reducing the amount of data contaminated by redundant information such as numbers.

\begin{figure}[t]
    \begin{center}
    \end{center}
    \includegraphics[width=\linewidth]{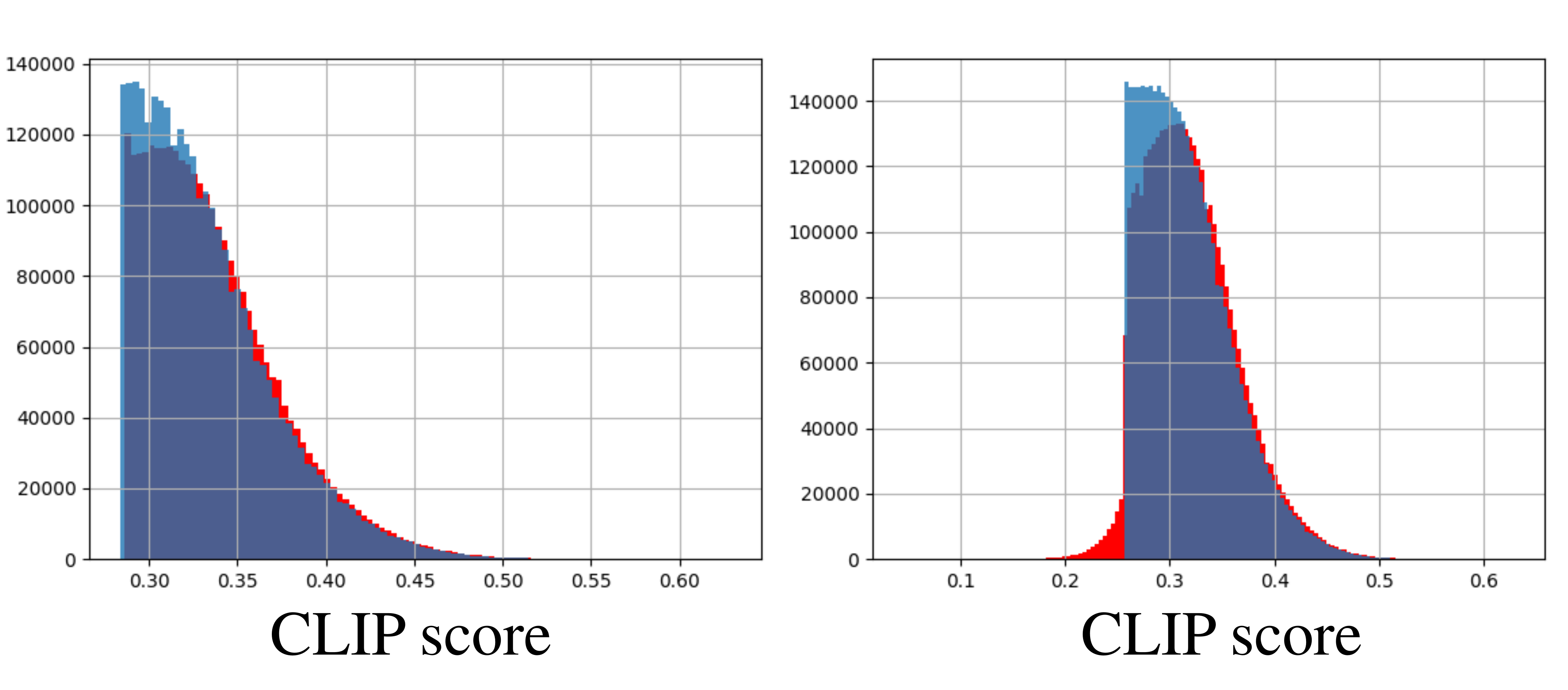}
    \caption{\textbf{Left}: The distribution of the data filtered by text-masked filter (top 30\%). \textbf{Right}:The distribution of the data filtered by text-masked filter (top 40\%). The red represents the CLIP score distribution of the filtered data, and the blue represents the CLIP score distribution of the same data where the data has been masked by our text-masker.}
    \label{fig:data-distribution}
\end{figure}

\vspace{-10pt}
\paragraph{Analysis of the distribution of CLIP scores after filtering.} Figure~\ref{fig:data-distribution} illustrates that our filter leads to a reduction in data with CLIP similarity scores below 0.32 in both cases. Notably, in the top 40\% case, we observe an influx of data with significantly lower CLIP scores. These findings are further elucidated by Table~\ref{tab:imagenet}, where the text-masked CLIP 40\% dataset outperforms the CLIP 40\% dataset. However, when filtering out relatively low CLIP score data from the text-masked CLIP 40\% dataset, the resulting performance is the lowest. Consequently, the increase of data with lower CLIP scores, rather than the loss of data with higher CLIP scores, improves the performance on ImageNet-1k.

\begin{table}[t]
    \centering
    \resizebox{0.8\linewidth}{!}{
    \begin{tabular}{@{}l|c@{}}
    \toprule
      Filter & ImageNet-1k Accuracy\\
    \midrule
         CLIP 30\%&0.0503\\
         text-masked CLIP 30\%&0.0483\\
         \midrule
         CLIP 40\%& 0.0448\\
         text-masked CLIP 40\%& 0.0464\\
         TMC 40\% threshold 0.0255& 0.0442\\
         \midrule
         CLIP 50\% & 0.0419\\
         text-masked CLIP 50\%&0.0439\\
    \bottomrule
    \end{tabular}
    }
     \caption{
     The CLIP score for each dataset is presented. The TMC 40\% threshold of 0.0255 indicates that data with a CLIP score lower than 0.0255 in the text-masked 40\% dataset should be filtered out. In the right-hand portion of Figure~\ref{fig:data-distribution}, this corresponds to trimming the left tail of the red section, thereby filtering out the low CLIP score data that our filter does not capture.}
    \label{tab:imagenet}
\end{table}

\vspace{-10pt}
\paragraph{Analysis of the results on ImageNet-1k dataset.} The performance of our filter compared to the CLIP filter depends on the percentage of data being filtered.In Table~\ref{tab:imagenet} When filtering out 70\% of the data, the CLIP filter outperforms our filter. However, when filtering out 60\% or 50\% of the data, our filter performs better than the CLIP filter.
Notably, as the fraction of filtered data increases, the negative impact of this loss is mitigated through the enhancement of high-quality data and the reduction of low-quality data.

\begin{table}[t]
  \renewcommand{\arraystretch}{1.1}
  \small
  \centering
  
  \resizebox{\linewidth}{!}{
  \begin{tabular}{@{}l|cccc@{}}
  \toprule
    Filter & size &  ImageNet & dist. shifts & VTAB   \\
  \midrule
   No filtering 
   & 12.8M& 02.5 & 03.3 & 14.5  \\
  Basic Filtering  
  & 3.0M  & 03.0 & 04.0 & 14.9  \\
   LAION filtering 
   & 1.3M  & 03.1 & 04.0 & 13.6 \\
   CLIP score (L/14 30\%) 
   & 3.8M  & \underline{05.1} & 05.5 & \underline{19.0} \\
   CLIP score (L/14 40\%) 
   & 5.1M  & 04.4 & 05.0 & 17.3  \\
   \midrule
    text-masked CLIP 30\% 
   & 3.8M  & 04.8 & \underline{05.7} & 18.8 \\
   text-masked CLIP 40\% 
   & 5.1M  & 04.6 & 05.1 & 17.1  \\
   
  \bottomrule
  \end{tabular}}
    \caption{Zero-shot accuracies for various filtering strategies on the \texttt{small} pools of the DataComp benchmark. Our text-masked CLIP 30\%method  outperforms the state-of-art on DataComp by a margin of 3.6\% on the small scale (ImageNet dist.shifts).Also, text-masked filter outperforms the original CLIP score filter when selecting the top 40\% of the data.}
  \label{tab:datacomp}
  \end{table}

\begin{figure}
    \centering
    \includegraphics[width=\linewidth]{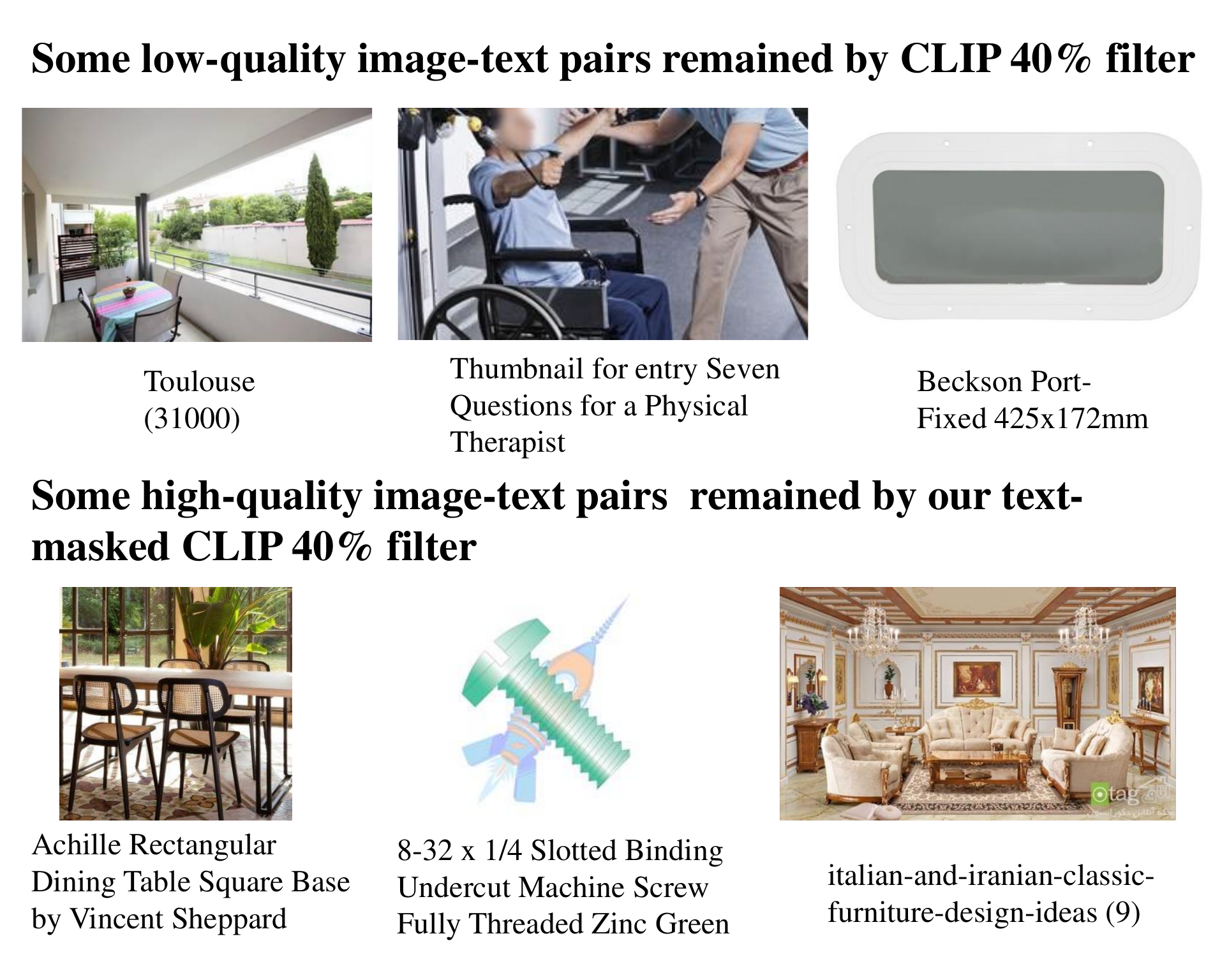}
    \caption{There are some examples screened respectively by CLIP 40\% filter provided by DataComp~\cite{gadre2023datacomp} and our text-masked CLIP 40\% filter.}
    \label{fig:examples}
\end{figure}

\section{Conclusion and Limitations}
In this article, we present a more robust text-masked filtering method that outperforms the CLIP score filter when applied at higher screening ratios. Through our text-masked filter, we introduce a subset of data contaminated by redundant information, resulting in lower CLIP scores. These data have a crucial impact on improving training effectiveness under specific screening ratios.

Furthermore, in our work, we also discovered the ambiguity of the CLIP model and the critical influence of redundant information on CLIP scores. The next step could involve investigating the correspondence between redundant information in text (such as numbers and bracketed contents) and specific elements in images. This insight can potentially inspire techniques for enhancing CLIP training through language rewrite.

It is important to note that our method may result in the loss of some high-quality data. Future research can explore strategies to prevent the loss of valuable data with high quality but low CLIP scores when incorporating them into the training process.

{
\small
\bibliographystyle{ieee_fullname}
\bibliography{main}
}

\end{document}